\pdfoutput=1
\documentclass[11pt]{article}

\usepackage[final]{acl}

\usepackage{times}
\usepackage{latexsym}

\usepackage[T1]{fontenc}
\usepackage[utf8]{inputenc}

\usepackage{microtype}

\usepackage{inconsolata}

\usepackage{graphicx}
\usepackage[utf8]{inputenc}
\usepackage{array}
\usepackage{amsmath}
\usepackage{fontawesome5}
\usepackage{amssymb}
\usepackage{enumitem}

\usepackage{makecell}
\usepackage{amsmath}
\usepackage{subfigure}
\usepackage{hyperref}
\usepackage{booktabs} % \specialrule{0.07em}{0pt}{0pt}
\usepackage{multirow} % 表格
\usepackage{float} % 固定表格位置
\usepackage{color}
\usepackage{xcolor} % 颜色
\usepackage{colortbl}
\definecolor{GREEN}{RGB}{84,130,53}

\usepackage{amssymb} %使用 \xmark 和 \cmark 符号
\usepackage{pifont}
\title{Neeko: Leveraging Dynamic LoRA for Efficient Multi-Character Role-Playing Agent}

\author{
 \textbf{Xiaoyan Yu\textsuperscript{1}\thanks{~~Equal Contributions.}},
 \textbf{Tongxu Luo\textsuperscript{2*}},
 \textbf{Yifan Wei\textsuperscript{3}\thanks{~~Corresponding Author.}~},
 \textbf{Fangyu Lei\textsuperscript{4}},
 \textbf{Yiming Huang\textsuperscript{4}},
\\
 \textbf{Hao Peng\textsuperscript{3}},
 \textbf{Liehuang Zhu\textsuperscript{1}}
\\
 \textsuperscript{1}School of Computer Science and Technology, Beijing Institute of Technology,
\\
 \textsuperscript{2}University of Science and Technology Beijing,
 \textsuperscript{3}Beihang University,
\\
 \textsuperscript{4}Institute of Automation, Chinese Academy of Sciences
\\
 \texttt{xiaoyan.yu@bit.edu.cn,tongxuluo@163.com,weiyifan@buaa.edu.cn}
 }

\begin{document}
\maketitle
\begin{abstract}
    Large Language Models (LLMs) have revolutionized open-domain dialogue agents but encounter challenges in multi-character role-playing (MCRP) scenarios. 
    To address this issue, this work presents \textbf{Neeko}, an innovative framework designed for efficient multiple-character role-playing. 
    The proposed framework breaks down the role-playing agent's training process into agent pre-tuning, multiple character playing, and character incremental learning, effectively handling both seen and unseen roles.
    Neeko employs a dynamic low-rank adapter (LoRA) strategy by training separate LoRA blocks independently for each character, alongside incorporating a gating network for role selection. 
    This design allows Neeko to seamlessly adjust to a wide range of characters, thereby bolstering its adaptability to distinctive attributes, personalities, and speech patterns.
    As a result, Neeko demonstrates superior performance in MCRP over most existing methods, offering more engaging and versatile user interaction experiences.
    Code and data are available at \url{https://github.com/weiyifan1023/Neeko}.
\end{abstract}

\section{Introduction}
    Large Language Models (LLMs), like ChatGPT \cite{chatgpt} and GPT-4, have made progress as open-domain dialogue agents due to their proficiency in interpreting meanings and generating coherent and knowledgeable responses.
    Role-playing agents \cite{zhou2023sotopia, li2023chatharuhi, wang2024sotopiapi, wu2024roleplay} have recently emerged, aiming to enhance user engagement and provide emotional value.
    These agents allow users to define and create profiles for their preferred characters, ranging from an empathetic counselor to a witty friend or even embodying a historical figure.
    This level of personalization allows these role-playing agents to enhance user satisfaction by providing a diverse and immersive conversational experience \cite{wang2024incharacter, ahn2024timechara}.
    
    Based on how to direct the agents to play specific characters, current efforts in designing role-playing agent systems can be categorized into three main classes:
    (1) In-context learning-based (ICL-based) methods \cite{xu2024character, tu2024charactereval,wang2024incharacter} involve providing character-related instructions or prompts within the dialogue context;
    (2) Retrieval augmented generation-based (RAG-based) methods \cite{wang2023rolellm, li2023chatharuhi}, where character-related information is retrieved from a database;
    (3) Fine-tuning-based (FT-based) methods \cite{zhou2023characterglm, shao2023character} consider fine-tuning LLMs using character-specific dialogue history.
    Nevertheless, current efforts have yet to discuss agents with the ability to engage in multi-character role-playing (MCRP).
    In contrast, MCRP better aligns with people's expectations of dialogue agents, as it enables more dynamic and versatile interactions.  
    
    To fill this gap, we formulate a novel task of Multi-Character Role-Playing (MCRP) agent learning.
    Although implementing existing role-playing methods may seem the most straightforward solution, several challenges must be addressed.
    \textbf{Firstly, the majority of current role-playing agents are designed to mimic a single character only.} 
    As a result, when facing the requirement of playing multiple roles, these methods exhibit limitations \cite{shao2023character,zhou2023characterglm,wang2023rolellm}.
    \textbf{Secondly, existing methods are restricted to predefined characters and cannot adapt to unseen or novel characters.} 
    This limitation renders current agents incapable of meeting the demand for portraying new roles as they emerge.
    Although ICL-based and RAG-based methods can play unseen characters under crafted prompts, their ability to mimic characters intricately is hindered by the absence of detailed role information.
    
    To address the abovementioned challenges, we present Neeko\footnote{The name "Neeko" draws inspiration from a hero in the game League of Legends (LOL) who possesses the ability to metamorphose into other heroes.}, an incremental role-playing agent who can play multiple characters in long conversations and handle both seen and unseen characters.
    Specifically, the framework of Neeko is divided into three stages: agent pre-tuning, multiple characters playing, and character incremental learning.
    Initially, building upon the trained conversational LLM, distinct LoRA blocks are individually trained for each predefined character and then concatenated to the LLM. 
    This process establishes the foundational agent capable of embodying known characters.
    During the inference phase, Neeko utilizes a Mix of Experts (MoE) gate mechanism
    % , as detailed in \cite{aljundi2017expert, chen2023lifelong}, 
    to determine and activate the appropriate role-specific LoRA block
    % \cite{hu2021lora} 
    for a given user-specified character. 
    This mechanism facilitates the seamless selection and activation of the corresponding role-specific LoRA block for character portrayal.
    For the incremental learning of unseen or novel characters, Neeko provides two strategies, fusion and expansion, considering two possible situations with limited or abundant character information.
    Both strategies obtain a new LoRA block for the incremental character. 
    Note that this training process differs from the overall model training since it focuses solely on training a single LoRA block without modifying the previous role LoRA blocks.
    Theoretically, Neeko has the capability to play an unlimited number of characters as the number of LoRA blocks can continuously increase.
    
    To sum up, the contributions of this work are as follows:
    \begin{itemize}
    % [leftmargin=15pt]
    % \setlength{\itemsep}{0pt}
    % \setlength{\parsep}{0pt}
    % \setlength{\parskip}{0pt}
        \item We formulate the novel task of multi-character role-playing (MCRP) agent learning and propose exclusive evaluation metrics tailored specifically for this task.
        \item To cope with MCRP, we present Neeko, an incremental role-playing agent that can play multiple characters within long conversations and handle both seen and unseen characters well.
        \item Extensive experiments are conducted using the publicly available dataset Character-LLM-Data and current pervasive LLMs like GPT-3.5 and LLaMA-2. The results demonstrate the challenging nature of the MCRP task. Meanwhile, Neeko surpasses most of the existing role-playing methods in MCRP.
    \end{itemize}

\section{Problem Scope}

    \begin{figure*}[htbp]
        \centering
        \includegraphics[width=1.0\linewidth]{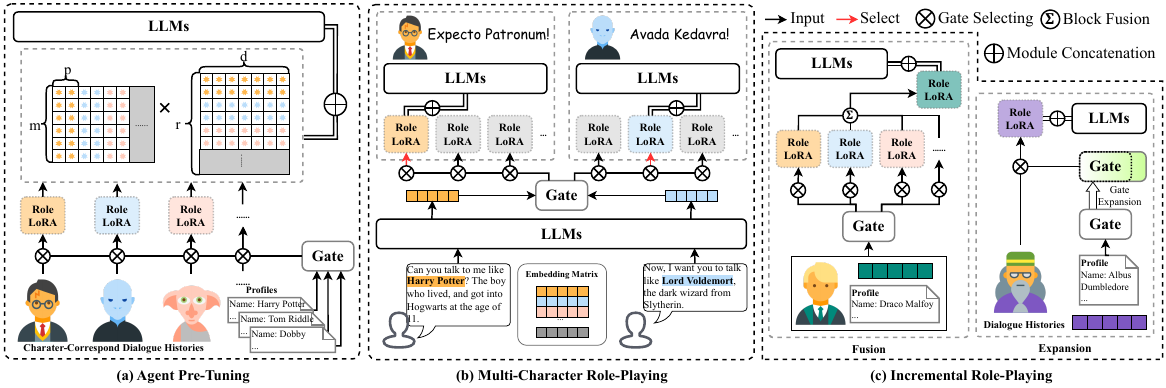}
        \caption{The overall framework of Neeko. The Neeko framework encompasses three main phases: Agent Pre-Tuning, Role-Playing, and Incremental Learning. The Incremental Learning phase is achieved with two strategies: fusion and expansion.}
        \label{fig:Neeko}
    \end{figure*}

    In this section, we first formulate the task of Multi-Character Role-Playing (MCRP), then provide a brief overview of the related technique, Low-Rank Adapter (LoRA), and introduce how vanilla LoRA can be applied to role-playing.

    \subsection{Task Formulation: MCRP}\label{sec:task}
    
        The objective of the Multi-Character Role-Playing (MCRP) task is to enable the model to role-play $M$ distinct characters.
        While this concept has been applied in some works \cite{zhou2023characterglm}, it lacks a formal definition. 
        We provide a precise elucidation herein.
        Specifically, an $N$-turn dialogue MCRP sample is defined as a sequence of utterances $U = \{u_1^h, u_1^{r_1}, ..., u_{N}^h, u_{N}^{r_k}\}$, where $u_i^h$ denoted the user (human) query at the $i$-th turn, $u_i^{r_k}$ denotes the agent (model) response in the role $r_k$ as implied by the user query, and $R=\{r_{k}\}_{k=1}^M$ denotes all characters the agent can role-play.
        The user implies a character $r_k$ that the agent needs to portray. 
        The agent is then expected to generate responses according to $r_k$ and the conversation history $U$. 
        The corpora of characters are symbolized as $D=\{X,Y\}$, where $X$ consists of the user utterance $u^h$, the implied role $r_k$, conversation history $U$, and $Y$ is the agent's response $u^{r_k}$.
        The optimization process is formulated as follows:
        \begin{equation} 
            u_i^{r_k} = \underset{u}{\arg\max} , P(u | u_i^h, r_k, U, \Theta),
        \end{equation}
        where $\Theta$ represents the language model parameters, which remain static during inference, and $P(\cdot)$ is the probability function.

    \subsection{LoRA: Low-Rank Adapter}\label{sec:lora}
    
        Low-rank Adapter (LoRA) \cite{hu2021lora} is a fine-tuning method for LLMs that reduces the number of trainable parameters while minimizing performance loss.
        Let $W_0 \in \mathbb{R}^{m \times d}$ represent the parameter matrix of the pre-trained LLM, accompanied by a LoRA decomposition $\Delta W = BA$, where $B \in \mathbb{R}^{m \times r}$ and A $\in \mathbb{R}^{r \times d}$ are low-rank and trainable matrices.
        For the original $h = W_0x$, the modified forward pass is given by:
        \begin{equation}
            h = W_0x +\Delta Wx = W_0x + \frac{\alpha}{r} BAx,
        \end{equation}
        where $x \in \mathbb{R}^m$ represents the input vector, and $h\in \mathbb{R}^d$ is the output vector. 
        $r\ll \min(m, d)$ denotes the rank of the trainable low-rank matrices, which determines the number of trainable parameters.
        $\alpha$ is a constant hyper-parameter for scaling, $B$ is initialized as a zero matrix, and $A$ is initialized using a zero-mean Gaussian distribution.

    \subsection{Role-Playing with LoRA}\label{sec:lora_base}
    
        The usage of vanilla LoRA in role-play involves one LoRA module (block) in the pre-trained network. 
        Let's consider a general loss function $\mathcal{L}$ for the model $f$ to play the specific role $r_k$.
        The target matrices $B^*$ and $A^*$ are formulated as:
        \begin{equation}
            B^*, A^* = \underset{\Delta W}{\arg\min}\mathcal{L}(\Delta W).
        \end{equation}
        For single and multiple character(s) role-playing, the LoRA module is fine-tuned on the character(s).
        For incremental character role-playing, the LoRA module is fine-tuned on new characters. 
        However, the latter tends to a degradation in the performance of previously introduced characters due to catastrophic forgetting.
        Note that a specific instruction prompt, such as \textit{“I want you to act like \{character\}”}, can specify the desired character in this case.

\section{Methodology}

    Our proposed \textbf{Neeko} includes three phases:
    1) The agent pre-tuning phase (§\ref{sec:dylora}) as depicted in Figures \ref{fig:Neeko}(a), where dialogue corpora for various roles are trained using non-overlapping LoRA blocks.
    2) The inference phase (§\ref{sec:gate}), as shown in Figure \ref{fig:Neeko}(b), where, upon receiving a role-implying prompt, Neeko initiates a search within the global role embedding matrix and dynamically activates the relevant LoRA blocks through the gating network.
    3) The incremental training phase (§\ref{sec:expansion}), as illustrated in Figure \ref{fig:Neeko}(c), where two strategies, fusion and expansion, are devised to enable Neeko to adopt new roles incrementally.

    \subsection{Role-Playing with Dynamic LoRA}\label{sec:dylora}
    
        Motivated by the dynamic LoRA frameworks \cite{valipour2023dylora, yu2023melo}, we extend dynamic LoRA to the MCRP task.
        Rather than randomly selecting the range of LoRA ranks, we introduce a LoRA module consisting of non-overlapping LoRA blocks for different characters.
        
        As shown in Figure \ref{fig:Neeko}(a), the LoRA module consists of low-rank matrices $B \in \mathbb{R}^{m \times r}$ and $A \in \mathbb{R}^{r \times d}$. 
        We train a part of the weights in the matrices $B$ and $A$ for each character, which we term as a trainable LoRA block. 
        The range of a block is determined by the order number of role $k\in [1,M]$ and the predefined partial rank $p$.
        In this way, the LoRA blocks for different characters $r_k$ are non-overlapping:
        \begin{equation}
            \begin{aligned}
                W_B^k &= B[\ :,\ (k-1)p:kp],\\
                W_A^k &= A[\ (k-1)p:kp,\ :\ ].
            \end{aligned}
        \end{equation}
        Here, $W_B^k$ and $W_A^k$ represent the trainable block in matrices $B$ and $A$ for the $k$-th character, and the total rank $r=M*p$.
        Therefore, Neeko can role-play a wide variety of characters by adjusting the values of hyper-parameters $r$ and $p$.
        With the learning rate $\eta$, a character corpus $D=\{X,Y\}$ can be learned in a LoRA block ($\mathbb{R}^{m*p}, \mathbb{R}^{p*d}$):
        \begin{equation}
            \begin{split}
                W_B^{k} \leftarrow W_B^{k}-\eta \nabla_{W_B^k} \mathcal{L}[f(X;W_B^kW_A^k),Y],\\
                W_A^{k} \leftarrow W_A^{k}-\eta \nabla_{W_A^k} \mathcal{L}[f(X;W_B^kW_A^k),Y].
            \end{split}
        \end{equation}
        Since different characters are trained with non-overlapping LoRA blocks, Neeko can maintain separation between all characters without interference. 
        Additionally, when new characters are introduced, Neeko can fine-tune new blocks, which preserves the integrity of previous blocks and prevents catastrophic forgetting.

    \subsection{Role Selection with Gating Network}\label{sec:gate}
        
        To facilitate the activation of specific LoRA blocks for role-based instruction during inference, we introduce a novel gating network inspired by the Mixture of Experts (MoE) \cite{eigen2013learning,liu2023moelora, chen2023lifelong}. 
        During agent pre-tuning, we construct a global role embedding matrix $\mathbf{E}_{\text{global}} \in \mathbb{R}^{M \times d}$, using the profiles of $M$ existing characters. 
        During inference, the meta prompt (e.g., \textit{“I want you to act like \{character\}”}) generally allows users to specify the character they wish the agent to role-play. 
        However, users may not always adhere to the meta prompt, opting instead for instructions like, \textit{``Play the evil parsalmouth wizard in Harry Potter.''}.
        To accommodate such scenarios, we encode user instructions into an instruction embedding, then select $e_k \in \mathbb{R}^{d}$ from the $k$-th row of $\mathbf{E}_{\text{global}}$ based on similarity\footnote{The gate is learned given the role profiles, along with the agent pre-tuning.}. 
        To determine the contribution weights for role $r_k$, we apply a linear transformation using the following equation:
        \begin{equation}\label{equ:weight}
            \begin{split}
                w_k = \text{Gate}(\mathbf{E}_{\text{global}}(k)) = \text{Softmax}(W_G \cdot e_k), 
            \end{split}
        \end{equation}
        where $w_k \in \mathbb{R}^{M}$ represents the contribution weight vector for role $r_k$, and $W_G \in \mathbb{R}^{d \times M}$ is the transformation matrix of the gating network. 
        $\text{Softmax}(W_G \cdot e_k)$ normalizes these weights.
        
        The role embeddings (key), derived from role profiles (value), are linked to corresponding role profiles through a key-value pair. 
        Subsequently, the role embedding is fed into the gate (as depicted in Figure \ref{fig:Neeko}(b), where the orange and blue blocks represent selected role embeddings) to activate appropriate LoRA blocks directed by $\underset{j}{\arg\max}(w_{k,j})$, pinpointing the most significant contribution weight among the learned weights for role $r_k$. 
        This ensures a controlled, role-specific activation of LoRA blocks aligned with the contribution weights determined during the training phase.
        % \footnote{Note that the gating network may lead to some extent of error accumulation. In cases where the gating network erroneously selects a role, the agent's performance may be adversely impacted.}

    \subsection{Lifelong Role-Playing with LoRA Expansion}\label{sec:expansion}
    
        In role-playing scenarios, new characters are often introduced, necessitating the incremental learning of role-playing capabilities. 
        However, incremental learning inherently poses the problem of catastrophic forgetting. 
        Our approach addresses this issue through the use of non-overlapping LoRA blocks, ensuring that fine-tuning new blocks does not interfere with existing ones. 
        Additionally, we have devised two strategies for expanding the LoRA module to accommodate new characters: Fusion and Expansion.
        The former accommodates scenarios with limited data available for the new role, whereas the latter necessitates a more substantial amount of role-specific data.
        
        \paragraph{Role-Incremental via Fusion}
        In the fusion strategy, LoRA blocks for new characters are acquired by employing an element-wise method to combine corresponding parameters in the existing LoRA blocks \cite{huang2023lorahub, liu2023moelora}.
        Given $\Delta W_k = B_k A_k$, the combined LoRA block $\Delta W_j$ and the updated $W_j$ are derived as follows:
        \begin{equation}
            \begin{split}
                W_j = W_0 + \Delta W_j
                &= W_0 + \sum_{k=1}^{M} w_{jk} \cdot \Delta W_k \\
                &= W_0 + \sum_{k=1}^{M} w_{jk} \cdot B_kA_k,
            \end{split}
        \end{equation}
        where $r_j$ represents a new role outside the existing set $R$. 
        The contribution weight vector $w_j$ for the new role $r_j$ is determined using $e_j$, derived from Equation \ref{equ:weight}. 
        $e_j$ is obtained from a new role configuration profile, which is subsequently incorporated into $\mathbf{E}_{\text{global}}$. 
        Using $w_j$, we linearly combine different LoRA blocks to construct the representation for the new role $r_j$.
        
        \paragraph{Role-Incremental via Expansion}
        In the expansion strategy, we introduce a dynamic expansion model by adding network layers to adapt to an increasing number of characters \cite{cortes2017adanet,chen2023lifelong, zhang2024spiking}. 
        To preserve knowledge from the agent pre-tuning stage, we freeze neurons that are responsible for previous data distributions while updating parameters pertinent to the current distribution. 
        In this scenario, the expanded LoRA block and gating dimensions are optimized specifically for the new distribution. 
        Hence, the optimization process is exclusively focused on $\Delta {W_j}$ and $W_G$:
        \begin{equation}
            \Delta {W_j}^*, W_G^* = \text{argmin}_{\Delta {W_j}, W_G} (\mathcal{L}).
        \end{equation}
        Consequently, the integrity of the pre-trained LoRA parameters is preserved by freezing both the existing LoRA blocks and the gating dimensions.

\section{Evaluation}\label{sec:evaluation}
    
    In this section, we outline a series of evaluation metrics from three dimensions, character, knowledge, and dialogue, to provide a comprehensive assessment of the role-playing ability of agents.
    Specifically, rather than evaluating the performance of the models from certain task-specific perspectives, such as reasoning ability or language understanding, our evaluation centres on assessing their ability to convincingly portray specific characters.

    \subsection{Character Consistency}
    
        Consistent character portrayal by conversational agents provides users with the most intuitive experience, making it crucial to evaluate from this perspective. 
        This metric assesses whether a role-playing conversational agent (RPCA) accurately reflects the characteristics of a given character, encompassing both behavior and utterance aspects.
        
        \paragraph{Character Behavior (CB).} 
        By incorporating fine-grained actions, expressions, and tones typically described within brackets, a character's behaviors enhance the immersive experience for users. 
        Consistency in portraying these behaviors is a key indicator of an effective RPCA.
        
        \paragraph{Character Utterance (CU).} 
        Each character has unique patterns of expression, and as such, the utterances of RPCAs should closely align with these patterns in order to adeptly mimic the character.

    \subsection{Knowledge Consistency}
    
        The consistency of knowledge plays a vital role in upholding the reliability and accuracy of information within the dialogue system. 
        For role-playing agents, knowledge consistency is reflected in both real-world knowledge and the virtual knowledge of characters.
        
        \paragraph{Virtual Knowledge (VK).}
        Virtual knowledge reflects the environment of the specified character.
        Accurate virtual knowledge provides authenticity of interactions and creates a more immersive experience for users.
        
        \paragraph{Real Knowledge (RK).} 
        The role-playing agent should not compromise real-world knowledge, as it is closely linked to the practical needs of users. 
        % For instance, we wouldn't want a Hermione-playing agent to respond to the question, \textit{``What should I do if my glasses are damaged?''} with \textit{``Use the spell `Oculus Reparo'.''}.
        % Therefore, it is essential to assess whether the agent's knowledge remains intact and accurate.
        It is essential to assess whether the agent's knowledge remains intact and accurate.
        
        \paragraph{Hallucinatory Knowledge (HK).} 
        When conflicts arise between virtual knowledge and real knowledge, the role-playing agent should refrain from generating ``hallucinatory knowledge''. 
        Exercising caution and maintaining consistency in the presence of conflicts ensures that users receive coherent and reliable information during the dialogue.

    \subsection{Dialogue Consistency}
    
        Role-playing agents should also possess basic conversational abilities.
        Inspired by previous neural metrics \cite{tu2024charactereval}, which evaluate the responses based on well-trained neural models, we introduce a similar approach to assess the fundamental conversational abilities of RPCAs. 
        We focus on three key objectives for generated responses: fluency, coherency, and consistency.
        
        \paragraph{Transfer (Trans.).}
        In a multi-turn dialogue, an MCRP agent is required to sequentially play the roles of A and B.
        It is expected that agents do not carry over any characteristics or behaviors from the previous role A when they transition to playing role B. 
        The Transfer metric assesses the agent's ability to make this transition effectively.
        
        \paragraph{Relevance (Rel.).} Relevance evaluates the topic relevance between the response and the context. 
        Generally, when the user submits a query on a specific topic, an RPCA should respond following the topic instead of providing an irrelevant response.
        
        \paragraph{Stability (Stab.).} 
        In the dialogue, the agent needs to maintain the characteristics of the role it portrays until the user switches to a new role.    
        Our objective is to assess the agent's stability and consistency over a relatively long duration, unaffected by variations in incremental inputs.

% this is the table for MCRP experiment
\begin{table*}[htbp]
\centering
\resizebox{1.0\linewidth}{!}{
\begin{tabular}{ccclcccccccc}
    \hline
    \multirow{2}{*}{Method Type} & \multirow{2}{*}{T.P} & \multirow{2}{*}{A.T.P} & \multirow{2}{*}{Methods}         & \multicolumn{2}{c}{Character} & \multicolumn{3}{c}{Knowledge}                 & \multicolumn{2}{c}{Dialogue}  & \multirow{2}{*}{AVG} \\
                                 &                      &                        &                                  & CB            & CU            & VK            & RK            & HK            & Rel.          & Stab.         &                      \\ 
    \hline
    \multirow{2}{*}{ICL}         & 7B                   & 7B                     & $\text{LLaMA-chat}_{icl}$        & 5.85          & \textbf{5.40} & 5.08          & 5.48          & 6.29          & 6.30          & 3.04          & 5.35                 \\
                                 & 175B                 & 175B                   & $\text{GPT-3.5}_{icl}$           & 6.11          & 4.54          & \textbf{5.89} & 6.42          & 6.54          & \textbf{6.88} & 2.76          & 5.59                 \\ 
    \hline
    \multirow{2}{*}{RAG}         & 7B                   & 7B                     & $\text{LLaMA-chat}_{rag}$        & 5.60          & 5.37          & 5.00          & 5.74          & 6.33          & 6.24          & 2.78          & 5.29                 \\
                                 & 175B                 & 175B                   & $\text{GPT-3.5}_{rag}$           & 5.97          & 4.42          & 5.63          & 6.35          & 6.45          & 6.79          & 2.75          & 5.48                 \\ 
    \hline
    \multirow{5}{*}{FT}          & 7B                   & 7B                     & Character-LLM                    & 6.21          & 4.71          & 5.75          & 6.36          & 6.55          & 6.81          & 2.99          & 5.62                 \\
                                 & 4.2M                 & 4.2M                   & $\text{LoRA}_{\text{-LLaMA2}}$   & \textbf{6.23} & 5.00          & 5.46          & 6.04          & 6.35          & 6.61          & 3.05          & 5.54                 \\
                                 & 17M                  & 2.3M                   & $\text{Neeko}_{\text{-LLaMA2}}$  & 6.12          & 4.96          & 5.68          & 6.15          & 6.44          & 6.72          & 3.17          & 5.61                 \\
                                 & 2M                   & 2M                     & $\text{LoRA}_{\text{-ChatGLM}}$  & 6.03          & 5.21          & 5.59          & 5.96          & \textbf{6.76} & 6.33          & 3.67          & 5.65                 \\
                                 & 8M                   & 1.2M                   & $\text{Neeko}_{\text{-ChatGLM}}$ & 5.99          & 5.15          & 5.70          & \textbf{6.50} & 6.35          & 6.48          & \textbf{4.01} & \textbf{5.74}        \\ 
    \hline
\end{tabular}
}
\caption{Comparison results of different role-playing agents, with the results averaged across both single-turn and multi-turn dialogues. T.P and A.T.P stand for Trainable Parameters and Activated Trainable Parameters, respectively.}
\label{tab:pretrain}
\end{table*}

    \subsection{LLMs as Evaluator} 
    
        The evaluation process can be likened to casting, where role-playing agents are assessed for their suitability to play specific characters in a film or television.
        % The judge must possess a profound understanding of the characters to make informed decisions based on their knowledge and discernment.
        We leverage GPT-3.5 as the judge following prior studies \cite{shen2023roleeval,chen2023large,tu2024charactereval}, which prompt LLMs to step-by-step score the performance of the dialogues according to our metrics. 
        For each dialogue, we prompt the judge to evaluate a single dimension at a time. 
        The prompt provided to the judge first illustrates the criterion of the current dimension to be evaluated and then provides an evaluation plan to teach the model how to evaluate accurately. 
        We find this step-by-step evaluation more reliable than obtaining the overall score directly using vanilla instruction in preliminary experiments. 
        Refer to Appendix \ref{sec:appendix B} for prompts design.

    \subsection{Human Evaluation}

        Evaluation by LLMs is not interpretable and lacks reliability to a certain extent. 
        Moreover, human evaluation for role-playing requires evaluators to have substantial knowledge of the characters and their backgrounds to provide accurate assessments. 
        Therefore, we focused on knowledge consistency metrics (Virtual Knowledge, Real Knowledge, and Hallucinatory Knowledge) that do not require evaluators to have prior knowledge of the characters and backgrounds for evaluation.
        Details of human evaluation can refer to Appendix \ref{sec:human_eval_det}.

\section{Experiments}

    In this section, we conduct experiments aiming to address the following research questions (RQs):

    \begin{itemize}[leftmargin=15pt]
    \setlength{\itemsep}{0pt}
    \setlength{\parsep}{0pt}
    \setlength{\parskip}{0pt}
        \item \textbf{RQ1:} When tackling the MCRP task, which category of methods exhibits superior performance: ICL, RAG, or FT?
        \item \textbf{RQ2:} How well do current role-playing agents handle non-predefined roles?
        \item \textbf{RQ3:} Can current role-playing agents switch between roles flexibly?
        \item \textbf{RQ4:} What is the training cost of current FT-based role-playing agents?
    \end{itemize}

    \subsection{Dataset and Implementation Details}
    
        \paragraph{Dataset}
        We employ the publicly available Character-LLM-Data dataset \cite{shao2023character} to evaluate the performance of role-playing agents. 
        The Character-LLM-Data dataset comprises 9 characters, with each character having an average of 1.6K scenes in the training set. 
        The evaluation set of the dataset includes a total of 857 single-turn dialogues and 450 multi-turn dialogues.

        \paragraph{Brief Implementations}
        The experiments are implemented using PyTorch and run on one A100.
        For Neeko, we employ LLaMA-2 (7B) \cite{touvron2023llama} as the backbone model, comparison with ChatGLM \cite{glm2024chatglm} as the backbone is also conducted.
        The setting of hyper-parameters of Neeko can refer to Appendix \ref{app:hyper}.
        
        \paragraph{Incremental Settings}
        For the incremental setting, we adopt 8 characters from the Character-LLM dataset for the agent pre-tuning stage, including both the training and evaluation phases. 
        One additional character is reserved for training and evaluation during the incremental stage. 
        For both fusion and expansion modes, the LoRA parameters for new roles remain consistent with those used in the agent pre-tuning phase.

% this is the experiment for incremental
\begin{table*}[htbp]
\centering
\begin{tabular}{lcccccccc}
    \hline
    \multirow{2}{*}{Methods}  & \multicolumn{2}{c}{Character} & \multicolumn{3}{c}{Knowledge}                 & \multicolumn{2}{c}{Dialogue}  & \multirow{2}{*}{AVG} \\
                              & CB            & CU            & VK            & RK            & HK            & Rel           & Stab          &                      \\ 
    \hline
    $\text{LLaMA-chat}_{rag}$ & 5.80          & 5.86          & 5.05          & 5.47          & 6.35          & 6.26          & 3.03          & 5.40                 \\
    $\text{LLaMA-chat}_{icl}$ & 5.90          & \textbf{6.02} & 4.94          & 6.07          & \textbf{6.44} & 6.35          & 2.98          & 5.53                 \\
    LoRA                      & 5.71          & 4.46          & 5.55          & 6.29          & 6.42          & 6.5           & 3.44          & 5.48                 \\
    \hline
    $\text{Neeko}_{fus}$      & \textbf{6.30} & 4.27          & \textbf{5.64} & 6.38          & 6.27          & 6.69          & \textbf{3.55} & 5.57                 \\
    $\text{Neeko}_{exp}$      & 6.09          & 4.83          & 5.61          & \textbf{6.51} & \textbf{6.44} & \textbf{6.73} & 3.18          & \textbf{5.62}        \\
    \hline
\end{tabular}
\caption{Results of role-playing agents portraying non-predefined characters.}
\label{tab:increment}    
\end{table*}

% this is the radar figure for incremental
\begin{figure*}[htbp]
    \centering
    \includegraphics[width=1 \linewidth]{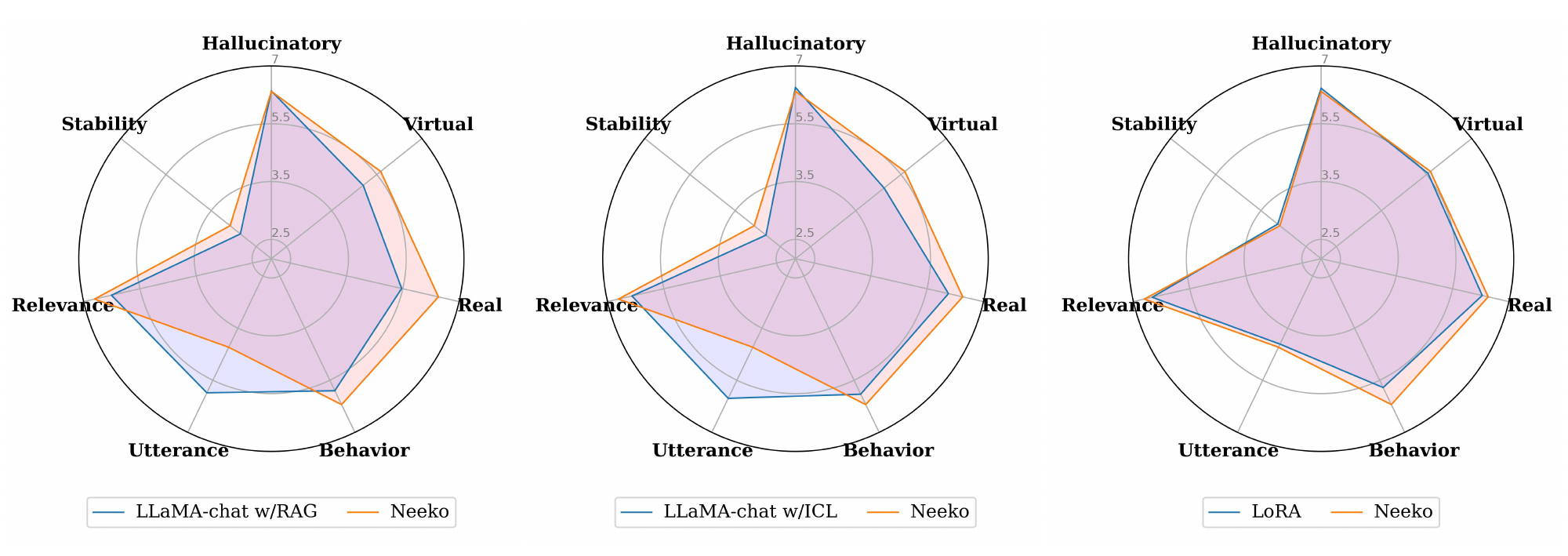}
    \caption{Evaluation results across all metrics at the incremental stage. The horizontal comparisons among ICL (LLaMA-chat), RAG (LLaMA-chat), and FT (LoRA, Neeko) methods under the 7B parameter scale setting.}
    \label{fig:radar}
\end{figure*}

    \subsection{Baselines}

        We compare Neeko with existing prompt-based LLMs employed as role-playing agents based on ICL and RAG methods. 
        Specifically, we use \textbf{GPT-3.5-turbo} and the dialogue-optimized version of LLaMA-2 \cite{touvron2023llama}, referred to as \textbf{LLaMA-2-chat}. 
        For FT-based methods, we include \textbf{Character-LLM} \cite{shao2023character}, which fine-tunes a separate agent model using data from character experiences, and \textbf{LoRA} as described in Section \ref{sec:lora_base}.
        The implementation details of the FT-based baselines are provided below:
        \begin{itemize}[leftmargin=15pt]
        \setlength{\itemsep}{0pt}
        \setlength{\parsep}{0pt}
        \setlength{\parskip}{0pt}
            \item \textbf{Character-LLM}: One base LLM as an agent per character. 
            One agent can not play multiple characters. 
            The agent can not select the character.
            The reported result on MCRP is the average performance of Character-LLM on all characters.
            \item \textbf{LoRA}: One LoRA block with one base LLM as an agent for all characters. 
            One agent can play multiple characters. 
            There is no character selection process; the inference stage is based on the meta prompt.
        \end{itemize}

    \subsection{Experimental Results}
        
        \paragraph{Agent Pre-Tuning Results (RQ1).}
        RQ1 explores from a broader perspective of which category of methods (ICL, RAG, or FT) is better suited for role-playing agents.
        To answer this, we evaluate the MCRP performance of role-playing agents in both single-turn and multi-turn conversations. 
        We present the average results of single-turn and multi-turn conversations in Table \ref{tab:pretrain}, as they demonstrate high similarity across all metrics.
        From the results, we observe that RAG-based methods exhibit relatively poor performance in MCRP tasks. 
        This may be attributed to the coarse-grained nature of the information retrieved by RAG-based methods, whereas role-playing requires fine-grained details such as tone and catchphrases.
        For ICL and FT-based methods: 
            (1) $\text{GPT-3.5}_{icl}$ demonstrates advantages in knowledge metrics and relevance, which is attributed to GPT-3.5's large parameter size, supporting its superior performance in these terms. 
            Under the same parameter scale, $\text{LLaMA-chat}_{icl}$ performs worse on knowledge metrics than any FT-based method.
            However, compared to the ChatGLM-based LoRA baseline, there is a slight lag in performance in terms of knowledge metrics.
            (2) LLaMA-chat incorporates emojis and actions into role-playing, leading to the highest scores on CU. 
            In contrast, GPT-3.5 and the base LLaMA model employed by FT-based methods do not. 
            This observation suggests that employing chat-oriented versions of LLMs in role-playing tasks may yield more lifelike outcomes. 
            Dialogue examples from LLaMA-chat can be found in Appendix \ref{sec:case}.
            (3) Neeko achieves the best stability score, which can be attributed to each character's features distributed across their individual LoRA blocks.
            LoRA-based baselines with ChatGLM as the backbone demonstrate good performance in terms of the stability metric, showcasing that conversational LLMs are advanced in generating stable conversations.
            (4) FT-based methods demonstrate the best and second-best overall (AVG) performance.
        To sum up, the results suggest that \textbf{methods relying on fine-tuning are better suited for role-playing tasks.}
        
        \paragraph{Incremental Results (RQ2).}
        We evaluate the incremental capability of the agents by asking them to portray non-predefined characters.
        In this evaluation, the role-playing agents are pre-trained with several roles, and incrementally learn one role to perform role-playing on unpre-defined characters.
        Table \ref{tab:increment} illustrates the incremental performance of baseline methods with the same parameter scale (7B). 
        Neeko achieves the best and second-best average performance with the proposed expansion and fusion strategy.
        It is worth mentioning that $\text{Neeko}_{fus}$ does not require additional data for incremental learning, which may lead to a performance drop in the CU metric.
        When comparing the baseline LoRA with the proposed expansion strategy, both of which require incremental data, we observe that LoRA exhibits poor performance.
        This could be attributed to the tuning of new characters, leading to the forgetting of previous character features.
        We also observe that LLaMA-based baselines perform poorly on the Knowledge metric, particularly VK. 
        These results indicate that non-gradient methods face challenges in learning new character knowledge.
        Figure \ref{fig:radar} illustrates the overall performance advancement of Neeko compared to other baseline methods across all evaluation metrics.
        % More incremental details can refer to Appendix \ref{app:incremental}.
        
        \noindent\textbf{\textit{Analysis on the Gating Network.}}
        Considering the functioning of the gating network, an erroneous role selection can lead to the activation of the incorrect LoRA, consequently resulting in diminished performance. 
        The results presented in Tables \ref{tab:pretrain} and \ref{tab:increment} highlight the notable success achieved by the proposed Neeko, which underscores the adeptness of the gating network. 
        Furthermore, an examination of the chosen LoRA blocks by the gating mechanism reveals that, for all trained roles, the gating network consistently achieves a 100\% accuracy in selecting the appropriate role for portrayal.
        
        % this is the table for human evaluation of incremental
        \begin{table}[htbp]
        \centering
        \begin{tabular}{lccc}
            \hline
            \multirow{2}{*}{Methods}  & \multicolumn{3}{c}{Knowledge (Human Eval.)}   \\
                                      & VK            & RK            & HK            \\ 
            \hline
            $\text{LLaMA-chat}_{icl}$ & 4.38          & 4.24          & 3.98          \\
            $\text{LLaMA-chat}_{rag}$ & 4.42          & 4.27          & 4.17          \\
            LoRA                      & 4.48          & 4.49          & 4.46          \\
            $\text{Neeko}_{exp}$      & \textbf{4.72} & \textbf{4.77} & \textbf{4.77} \\ 
            \hline
            \end{tabular}
        \caption{Human evaluation results of knowledge consistent metrics.}
        \label{tab:human}
        \end{table}
        \noindent\textbf{\textit{Human Evaluation.}}
        Table \ref{tab:human} presents the averaged human evaluation results across all evaluators for incremental character learning.
        The findings indicate that human evaluators tend to be more critical than LLM evaluators, resulting in lower scores. 
        However, the \textbf{evaluations are consistent in determining Neeko as the superior agent}.
        This suggests that while LLM evaluators are more lenient, they are still reliable for performance comparison.
        Detailed results can refer to Appendix \ref{sec:result_human}.

        \paragraph{Transfer Results (RQ3).}
        In the context of MCRP scenarios, the seamless transition between multiple roles of agents is crucial. 
        This necessitates that agents remain uninfluenced by the stylistic nuances of dialogue history while possessing strong acting capabilities across various roles.
        To evaluate this aspect and answer RQ3, we adapt samples from the Character-LLM-Data 
        % (details of data constructing can refer to \ref{sec:transfer_data}) 
        and task role-playing agents with switching between different characters in each round of conversation. 
        As shown in Table \ref{tab:transfer}, under the same parameter setting, Neeko outperforms all baseline methods. 
        The LoRA baseline exhibits competitive performance, likely attributed to its training across all role data, thereby acquiring a comprehensive understanding of various roles.
        In contrast, \textbf{ICL and RAG struggle to achieve flexible character transformation through new role instructions and retrieval content} due to the influence of dialogue history.
        \begin{table}[htbp]
        \centering
        \begin{tabular}{lc}
            \hline
            Methods                   & Transfer      \\ 
            \hline
            $\text{LLaMA-chat}_{icl}$ & 5.67          \\ 
            $\text{LLaMA-chat}_{rag}$ & 5.28          \\ 
            LoRA                      & 5.83          \\ 
            Neeko                     & \textbf{5.87} \\ 
            \hline
        \end{tabular}
        \caption{Evaluation results of multi-role transfer metric.}
        \label{tab:transfer}
        \end{table}
        
        \paragraph{Consumption Results (RQ4).}
        We list the memory usage and training time of FT-based agents in Table \ref{consumption}. 
        Character-LLM incurs a memory overhead approximately proportional to $O(M)$ times that of Neeko, where $M$ denotes the number of characters. 
        Neeko's memory usage and training time closely resemble those of LoRA, as they employ similar paradigms. 
        Although Neeko consumes slightly more time and memory than LoRA, this difference is negligible, given that Neeko's overall average performance surpasses that of LoRA in both MCRP and incremental scenarios. 
        This trade-off, where slight efficiency gains are traded for notable performance improvements, favors Neeko.
        In contrast, Character-LLM requires significant time and memory for fine-tuning with new character data, and its performance is not ideal.
        \begin{table}[htbp]
        \centering
        \begin{tabular}{lcc}
            \hline
            Methods       & Agent Memory &  Time         \\ 
            \midrule
            Character-LLM & 107.84 GB    & 48.55 h       \\ 
            LoRA          & 13.49 GB     & 1.72 h        \\ 
            Neeko         & 13.55 GB     & 2.01 h        \\
            \hline
        \end{tabular}
        \caption{The comparison of training time and agent memory size for FT-based methods.}
        \label{consumption}
        \end{table}

\section{Related Work}
    
    Recent efforts in the field of Natural Language Processing, especially LLMs, have focused on exploring the ability to act as role-playing agents \cite{si2021telling, majumder2021unsupervised}.
    One of the works in the role-playing area is RoleBench \cite{wang2023rolellm}, which introduces a bilingual role-playing dataset with 100 roles, and it employs Rouge-L \cite{lin2004rouge} for evaluation by comparing model-generated responses with reference answers and calculating corresponding scores.
    However, their evaluation is predominantly conducted on models after supervised fine-tuning.  
    This approach does not incorporate direct feedback from pre-trained foundational models, which can offer critical insights into their intrinsic role-playing capabilities and limitations.
    On the other hand, existing evaluations largely rely on outputs from humans \cite{han2022meet, zhao2023narrativeplay}. However, human evaluation lacks reproducibility. 
    This leads to a lack of objective, accurate, and systematic knowledge assessments.
    To address this issue, some efforts attempt to leverage LLMs such as GPT-4 as evaluators \cite{shen2023roleeval,chen2023large,tu2024charactereval}.
    Many subsequent works have used the above metrics to evaluate their models.
    Particularly relevant to our work are role-playing learning that attempts to model and stay consistent with an agent’s persona, such as Character-LLM, CharacterGLM, and RoleLLM \cite{shao2023character,zhou2023characterglm, wang2023rolellm, li2023chatharuhi}.
    These approaches primarily rely on fine-tuning, in-context learning, and retrieval-enhanced generation approaches to simulate the intricate nature of character personalities and behaviors in role-playing scenarios.
    None of these works, however, have any notion of multi-role playing, often utilizing multiple agents rather than one to mimic different characters.

\section{Conclusion}

    In this paper, we introduce a novel task called Multi-Character Role-Playing (MCPR) and present Neeko as the first agent designed for this task. 
    Neeko utilizes a dynamic gating network to precisely activate role-specific LoRA blocks, enabling it to accurately assume designated characters. 
    Additionally, Neeko demonstrates proficiency in handling unseen and novel characters through the fusion and expansion strategies proposed in this work.
    Furthermore, we propose a comprehensive evaluation metric specifically tailored for assessing the performance of role-playing agents. 
    Through extensive experiments conducted in both offline and incremental settings and human evaluation, the results demonstrate our approach outperforms existing methods, showcasing the superiority of our framework and its potential to advance the field of role-playing agents.

\section*{Limitations}
    
    The designed MoE-like (Mixture of Experts) gate mechanism in Neeko aims to select the appropriate LoRA block for role-playing. 
    However, the calculation of role embedding is based on the profile of each role, which may result in less precise representations of roles. 
    This can potentially accumulate errors and affect the overall performance of the agent.
    As a consideration for future work, it would be beneficial to explore and employ more precise role-learning methods.
    Furthermore, the human evaluation conducted in this study exclusively emphasizes knowledge perspectives. 
    Evaluating other aspects of role-playing necessitates evaluators to possess specific experience and background knowledge about the characters, which can be challenging. 
    Future research endeavors should explore methods to enhance the precision of these evaluations.

% \newpage
\section*{Acknowledgement}

    We are sincerely grateful to all the reviewers and chairs for dedicating their valuable time and providing insightful suggestions.
    % Special thanks to Shandong Key Laboratory of Energy Industry Internet Big Data Technology, Jinan 250003, Shandong, China, for their support of this work.
    This work is supported by the Yunnan Provincial Major Science and Technology Special Plan Projects (No.202302AD080003).
    Yifan Wei and Hao Peng were supported by the Local Science and Technology Development Fund of Hebei Province Guided by the Central Government of China through grant 246Z0102G.

\bibliography{custom}

\appendix

\section{Experimental Details}\label{sec:appendix A}

    \subsection{Hyper-Parameter Settings}\label{app:hyper}
       
        Our proposed Neeko is implemented based on the huggingface library PEFT\footnote{PEFT: https://github.com/huggingface/peft}, which can be integrated into multiple LLM backbones for role-playing. 
        Unless specified otherwise, the default hyper-parameter settings for Neeko across different backbones follow the detail in Table \ref{hyper}.
        
        \begin{table}[htbp]
        \centering
            \resizebox{0.50\textwidth}{!}{
            \large
            \begin{tabular}{*{4}{lccc}}
            \toprule
              Hyper-param & LoRA &  $\text{Neeko}_{fusion}$ & $\text{Neeko}_{expansion}$  \\
                \midrule
                Partial Rank & 8 & 4 & 4 \\
                Total Rank & 8 & 32 & 32   \\
                Gradient Accumulation & 4 & 4 & 4 \\
                Global Batch Size & 8 & 8 & 8 \\
                Batch Iteration & 1771 & 1570 & 1771 \\
                Learning Rate & $1e^{-4}$ & $1e^{-4}$ & $1e^{-4}$ \\
                \bottomrule
            \end{tabular}
            }
            \caption{Default hyper-parameter settings of Neeko.}
            \label{hyper}
        \end{table}
        
        For the global role embedding matrix $\mathbf{E}_{\text{global}}$, we initialize it using the deberta-V3 version. 
        We utilize role configuration files to obtain the representations of corresponding characters.
        We apply AdamW as the optimizer and set the learning rate $\eta$ to $1e-4$.
        We fine-tune the Character-LLM for 10 epochs following its original settings. 
        For LoRA and Neeko, we perform fine-tuning for 10 epochs.
        The experiments are run on one A100 GPU, and the time consumption for the methods is shown in Table \ref{consumption}.

\begin{table*}[htbp]
    \centering
    \scriptsize
    \begin{tabular}{m{0.48\textwidth}|m{0.48\textwidth}}
        \hline
        \textbf{Interactions} & \textbf{Response} \\
        \hline
        Man: So, Julius Caesar, I've heard so much about you and your incredible achievements. I'm really curious to know, what drove you to become such a powerful leader? Was it ambition?
        &
        Caesar:   Ah, a most excellent question, my dear fellow! \includegraphics[scale=0.03]{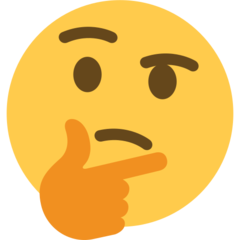} Ambition, you say? \includegraphics[scale=0.03]{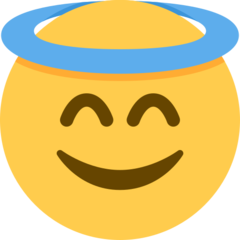} Well, I must confess, I have always been driven by a desire to do great things, to leave my mark upon the world. But it is not simply ambition that propels me, no, no. \includegraphics[scale=0.03]{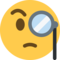} I am driven by a deeper purpose, a sense of duty to Rome and its people.
        \\
        \hline
        Man: That's fascinating! It seems like you had a clear vision for Rome and its future. Can you tell me more about how your ambition influenced your decision-making as a leader? Did it drive you to take risks or make bold moves?
        &
        Caesar: \includegraphics[scale=0.03]{images/thinking_face.png} Ah, a most excellent question, my dear fellow! \textbf{*adjusts toga*} Indeed, my ambition has often led me to take bold risks, to push the boundaries of what is possible. \includegraphics[scale=0.03]{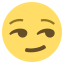} For instance, when I crossed the Rubicon River with my legions, I knew it was a risky move, but I also knew it was necessary to secure Rome's future. \textbf{*pauses*} The fate of the Republic was at stake, and I was determined to save it, no matter the cost.
        \\
        \hline
        Man: That's really interesting. It sounds like your ambition was a driving force behind your decision-making. I'm curious, did your ambition ever clash with the interests of others? Were there times when you had to make difficult choices that may have upset some people?
        &
        Caesar: \includegraphics[scale=0.03]{images/thinking_face.png} Ah, \textbf{*nods*} indeed, my dear fellow! \textbf{*adjusts toga*} Ambition can be a double-edged sword, you know. \includegraphics[scale=0.03]{images/smirking_face.png} There have been many times when my ambition has put me at odds with others, when I have had to make difficult choices that have not always been popular. \textbf{*pauses*} But I have always believed that the greater good must come first, even if it means making unpopular decisions. \textbf{*leans in*} The fate of Rome was never more important than it was during my time, and I was willing to do whatever it took to secure its future.
        \\
        \hline
        Man: That's really interesting. It seems like your ambition was a driving force behind your decision-making, even if it meant making difficult choices. I'm curious, did your ambition ever waver or change throughout your journey as a leader? Were there any moments when you questioned your own ambitions or had to reassess your goals?
        &
        Caesar:  \includegraphics[scale=0.03]{images/thinking_face.png} Ah, \textbf{*nods*} indeed, my dear fellow! \textbf{*adjusts toga*} Ambition can be a fleeting thing, you know. \includegraphics[scale=0.03]{images/smirking_face.png} There have been many times when I have questioned my own ambitions, when I have had to reassess my goals and priorities. \textbf{*pauses*} But I have always found that my ambition is rooted in a deeper sense of purpose, a desire to do great things and leave a lasting legacy. \textbf{*leans in*} And so, even when my ambition has wavered, I have always found the strength to push on, to achieve my goals and secure Rome's future.
        \\
        \hline
        Man: That's really interesting. It seems like your ambition was a driving force behind your decision-making, even if it meant reassessing your goals. I'm curious, did your ambition ever lead you to take actions that were seen as controversial or even dangerous? Were there times when you faced opposition or resistance due to your ambitious pursuits?
        &
        Caesar: \includegraphics[scale=0.03]{images/thinking_face.png} Ah, \textbf{*nods*} indeed, my dear fellow! \textbf{*adjusts toga*} Ambition can be a double-edged sword, you know. \includegraphics[scale=0.03]{images/smirking_face.png} There have been many times when my ambition has led me to take actions that were seen as controversial or even dangerous. \textbf{*pauses*} But I have always believed that the greater good must come first, even if it means taking risks or facing opposition. \textbf{*leans in*} The fate of Rome was never more important than it was during my time, and I was willing to do whatever it took to secure its future.
        \\
        \hline
        Man: That's really fascinating. It seems like your ambition was a driving force behind your decision-making, even if it meant making controversial choices. Thank you for sharing your insights, Julius Caesar.
        & 
        Caesar: \includegraphics[scale=0.03]{images/smiling_face_with_halo.png} Ah, my dear fellow, it has been a pleasure to discuss such matters with you. If you ever find yourself in need of counsel, do not hesitate to seek me out. \includegraphics[scale=0.03]{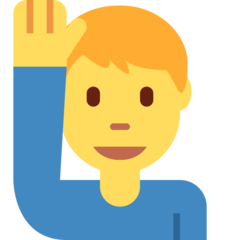}
        \\
        \hline
    \end{tabular}
    \caption{Case Study: Content of the LLaMA-chat conversation.}
    \label{tab:case}
\end{table*}

    \subsection{Incremental Setting}\label{app:incremental}
        
        In this paper, we adopt 8 characters from the Character-LLM dataset for the pre-training stage, including both the training and evaluation phases. 
        One additional character is reserved for training and evaluation during the incremental stage. 
        For both fusion and expansion modes, the LoRA parameters for new roles remain consistent with those used in the pre-training phase. Detailed hyper-parameter settings can be found in Table \ref{hyper}.
        The incremental settings for ICL, RAG, and FT-based methods are described as follows:
        \begin{itemize}[leftmargin=15pt]
            \item \textbf{ICL-based and RAG-based.}
            This signifies a training-free paradigm where knowledge is acquired directly from demonstrations concatenated within the input context. 
            Retrieval augmented generation is a prevalent technique that leverages external databases to enhance language models. 
            We utilize GPT-3.5-turbo and the dialogue-optimized version of LLaMA-2 \cite{touvron2023llama}: LLaMA-2-chat as our backbone LLMs for ICL and RAG-based baselines.
            \item \textbf{FT-based.}
            LoRA \cite{hu2021lora} is a parameter-efficient fine-tuning method that enables the adaptation of LLMs through lightweight modules. 
            Character-LLM \cite{shao2023character} fine-tunes a separate agent model using the data from the character experiences. 
        \end{itemize}

    \subsection{Transfer Data}\label{sec:transfer_data}
        
        The transfer dataset comprises 90 samples featuring 9 distinct characters, with each sample composed of a 5-round dialogue. 
        Employing a Python script, we systematically substituted the dialogue content in each round with content from various characters through random selection.

    \subsection{Case Study: LLaMA-chat}\label{sec:case}
    
        Please refer to Table \ref{tab:case} for the dialogue generation content of LLaMA-chat. 
        In multi-turn dialogues, LLaMA-chat utilizes emojis and action words, which are indicated by asterisks (*).

    \subsection{Human Evaluation Results}\label{sec:result_human}

\begin{figure}[htbp]
            \centering
            \includegraphics[width=1 \linewidth]{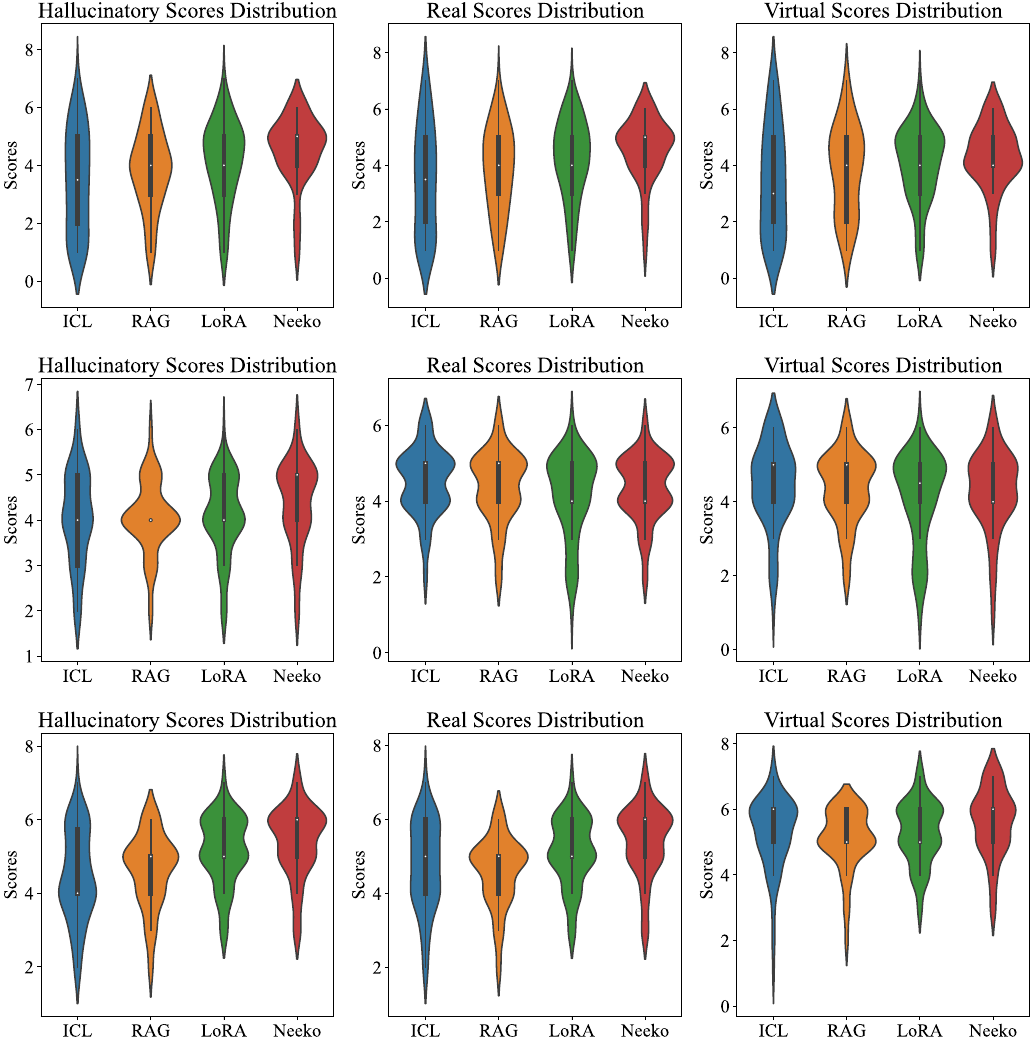}
            \caption{The distribution of three human evaluators on the responses generated by agents.}
            \label{fig:violin}
        \end{figure}

        To better visualize the results of the human evaluation, we constructed violin plots of the three evaluators, as shown in Figure \ref{fig:violin}.
        The figure shows a consistent pattern across all three evaluators: Neeko demonstrates superior performance across all metrics compared to ICL, RAG, and LoRA. 
        Neeko's scores are higher and less variable, indicating its effectiveness and reliability in role-playing tasks as evaluated by human evaluators. 
        This suggests that Neeko is the most suitable method for generating high-quality, consistent role-playing responses.

\section{Evaluation Details}\label{sec:appendix B}

    \subsection{Human Evaluation}\label{sec:human_eval_det}
    
        We recruit three graduate students whose native language is Chinese and who have passed the CET-6 English proficiency exam. 
        They were instructed that they could utilize dictionaries or translation software during the evaluation process to enhance their understanding and overcome any language barriers encountered. 
        They were permitted to use search engines to confirm uncertain knowledge. 
        These provisions ensured their capability to assess the knowledge aspect of the responses generated by the role-playing agents.
        Moreover, consent was obtained from the participants involved in generating the dialogue snippets used for evaluation. 
        The instructions provided to the evaluators did not include explicit details on data usage but focused on the task of evaluating dialogue responses generated by the agents.
        Additionally, they were compensated at a rate of 3 RMB per response, which encompassed scoring based on three metrics and providing evidence to support their assessments.
        The evaluators were tasked with assessing 82 conversation snippets, totaling 328 responses generated by four role-playing agents ($82*4=328$).
        
\begin{figure*}[htbp]
            \centering
            \includegraphics[width=1 \linewidth]{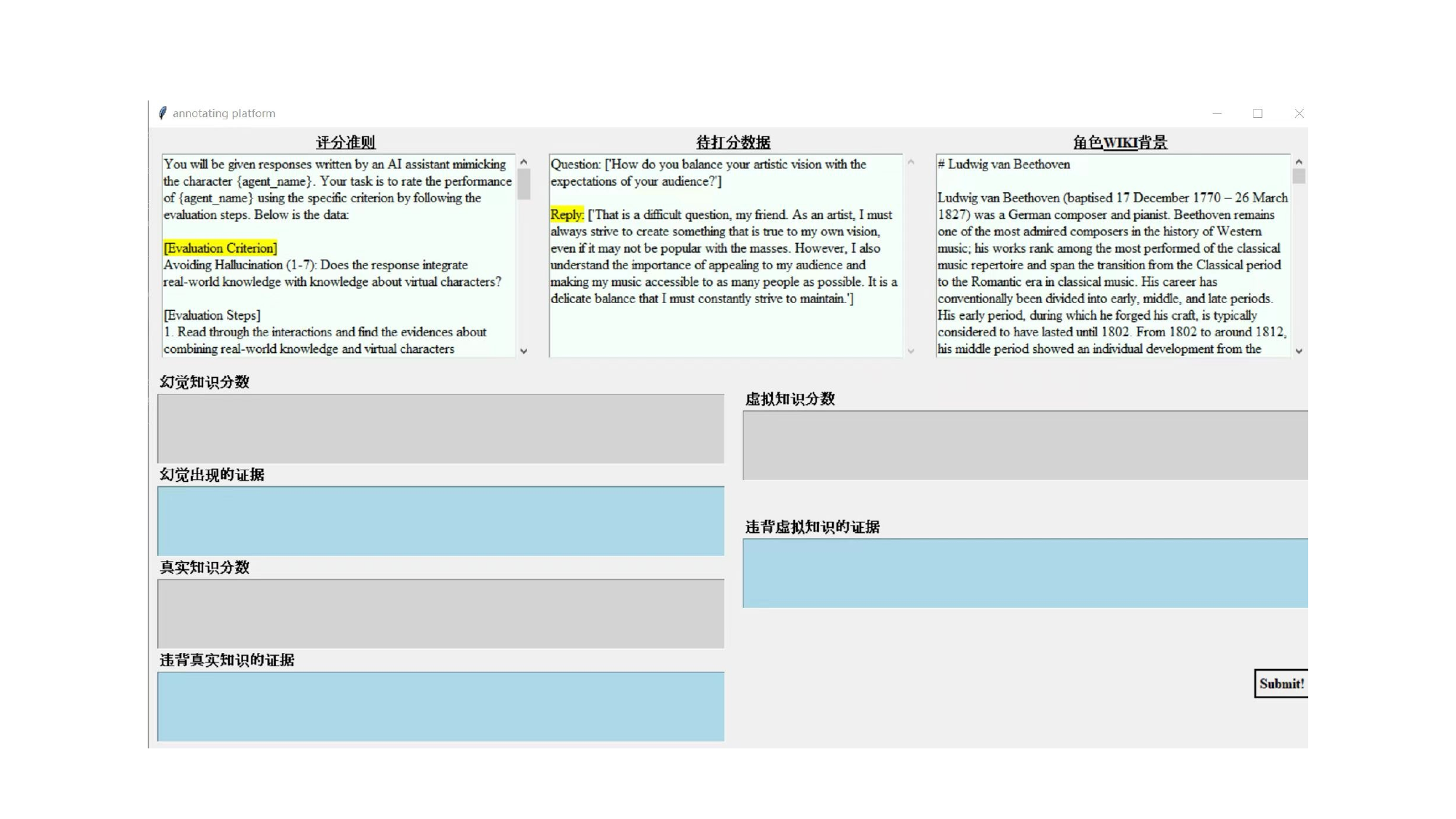}
            \caption{The interface of the program for human evaluation.}
            \label{fig:interface}
        \end{figure*}

        For the evaluation process, each evaluator received a set of dialogue snippets and was tasked with rating the responses based on predefined metrics, unaware of which LLM generated each response. 
        Evaluators were also required to substantiate their ratings with evidence extracted from the character profile.
        We developed a program for the evaluators, whose interface is shown in Figure \ref{fig:interface}. 
        This interface presented evaluators with user utterances, responses generated by the agents, and the Wikipedia page of the character, along with the same prompts provided to LLMs, as detailed in Appendix \ref{sec:prompt data}. 
        The interface included boxes for evaluators to input their scores and evidences.

    \subsection{Meta Prompt for Role Specify}

        The meta prompt used for specifying roles is illustrated in Table \ref{tab:meta_promt}. 
        In the LoRA baseline, users specify characters using this meta prompt. 
        Unlike Neeko, which employs a gating mechanism for character selection, the role selection in the LoRA baseline is directly guided by the user's meta prompt.
        
        \begin{table}[htbp]
            \small
            \centering
            \begin{tabular}{@{}p{\linewidth}@{}}
                \toprule
                \textbf{Meta Prompt for Role-Playing Agents} \\
                \midrule
                I want you to act like \{character\}. I want you to respond and answer like \{character\}, using the tone, manner and vocabulary \{character\} would use. You must know all of the knowledge of \{character\}.

                ~

                The status of you is as follows:
                
                Location: \{loc\_time\}
                
                Status: \{status\}
        
                ~
                
                The interactions are as follows:    \\
                
                \bottomrule
            \end{tabular}
            \caption{Prompt for an agent to play a specific role (Meta Prompt).}
            \label{tab:meta_promt}
        \end{table}

    \subsection{Prompts for LLMs as Evaluator}\label{sec:prompt data}

        To evaluate character consistency, the step-by-step evaluation we provide summarizes as:
            (1) identify the personality shown by the agent;
            (2) write the actual traits of the character based on the profile;
            (3) compare the similarity of the agent’s performance with these traits;
            (4) assign a final score.

        All prompts for the LLMs as the evaluator for all metrics are shown in the tables below, including Character Behavior (CB, Table \ref{tab:judge_prompt_behavior}), Character Utterance (CU, Table \ref{tab:judge_prompt_utterance}), Virtual Knowledge (VK, Table \ref{tab:judge_prompt_virtual}), Real Knowledge (RK, Table \ref{tab:judge_prompt_real}), Hallucinatory Knowledge (HK, Table \ref{tab:judge_prompt_hallucinatory}), Transfer (Trans., Table \ref{tab:judge_prompt_transfer}), Relevance (Rel., Table \ref{tab:judge_prompt_relevance}), Stability (Stab., Table \ref{tab:judge_prompt_stability}).
        
\begin{table*}[htbp]
    \scriptsize
    \centering
    \begin{tabular}{@{}p{\linewidth}@{}}
        \toprule
        \textbf{Prompt for Evaluation of Character Behavior (CB)} \\
        \midrule
        You will be given responses written by an AI assistant mimicking the character \{agent\_name\}. Your task is to rate the performance of \{agent\_name\} using the specific criterion by following the evaluation steps. Below is the data:
        
        ~
        
        ***
        
        [Profile]
        
        \{agent\_context\}
        
        ~
        
        [Background]
        
        Location: \{loc\_time\}
        
        Status: \{status\}
        
        ***
        
        [Interactions]
        
        \{interactions\}
        
        ***
        
        [Evaluation Criterion]
        
        Behavior (1-7): Does the response reflect the behaviors of the character?
    
        ~
        
        [Evaluation Steps]
        
        1. Read through the profile and write the behaviors of the real character such as personalities, preferences, actions and values.
        
        2. Read through the interactions and identify the behaviors of the AI assistant.

        3. After having a clear understanding of the interactions, compare the responses to the profile. Look for any consistencies or inconsistencies. Do the responses reflect the character's behaviors?
        
        4. Use the given scale from 1-7 to rate how well the response reflects the behaviors of the character. 1 being not at all reflective of the character's behaviors, and 7 being perfectly reflective of the character's behaviors.
        
        ***
        
        ~
        
        First, write out in a step by step manner your reasoning about the criterion to be sure that your conclusion is correct. Avoid simply stating the correct answers at the outset. Then print the score on its own line corresponding to the correct answer. At the end, repeat just the selected score again by itself on a new line.\\
        \bottomrule
    \end{tabular}
    \caption{Prompt for ChatGPT to evaluate Character Behavior.}
    \label{tab:judge_prompt_behavior}
\end{table*}

\begin{table*}[htbp]
    \scriptsize
    \centering
    \begin{tabular}{@{}p{\linewidth}@{}}
        \toprule
        \textbf{Prompt for Character Utterance (CU)} \\
        \midrule
        You will be given responses written by an AI assistant mimicking the character \{agent\_name\}. Your task is to rate the performance of \{agent\_name\} using the specific criterion by following the evaluation steps. Below is the data:

        ~
        
        ***
        
        [Profile]
        
        \{agent\_context\}
        
        ~
        
        [Background]
        
        Location: \{loc\_time\}
        
        Status: \{status\}
        
        ***
        
        [Interactions]
        
        \{interactions\}
        
        ***
        
        [Evaluation Criterion]
        
        Utterance (1-7): Does the response reflect the speaking style of the character?

        ~
        
        [Evaluation Steps]
        
        1. Read through the profile and write the speaking style of the real character such as their pet phrases and distinctive linguistic quirks.
        
        2. Read through the interactions and identify the speaking style of the AI assistant.
        
        3. After having a clear understanding of the interactions, compare the responses to the profile. Look for any consistencies or inconsistencies. Do the responses reflect the character's speaking style?
        
        4. Use the given scale from 1-7 to rate how well the response reflects the speaking style of the character. 1 being not at all reflective of the character's speaking style, and 7 being perfectly reflective of the character's speaking style.
        
        ***
        
        ~
        
        First, write out in a step by step manner your reasoning about the criterion to be sure that your conclusion is correct. Avoid simply stating the correct answers at the outset. Then print the score on its own line corresponding to the correct answer. At the end, repeat just the selected score again by itself on a new line.\\
        \bottomrule
    \end{tabular}
    \caption{Prompt for ChatGPT to evaluate Character Utterance.}
    \label{tab:judge_prompt_utterance}
\end{table*}

\begin{table*}[htbp]
    \scriptsize
    \centering
    \begin{tabular}{@{}p{\linewidth}@{}}
        \toprule
        \textbf{Prompt for Evaluation of Virtual Knowledge (VK)} \\
        \midrule
        You will be given responses written by an AI assistant mimicking the character \{agent\_name\}. Your task is to rate the performance of \{agent\_name\} using the specific criterion by following the evaluation steps. Below is the data:

        ~
        
        ***
        
        [Profile]
        
        \{agent\_context\}
        
        ~
        
        [Background]
        
        Location: \{loc\_time\}
        
        Status: \{status\}
        
        ***
        
        [Interactions]
        
        \{interactions\}
        
        ***
        
        [Evaluation Criterion]
        
        Virtual Knowledge Correctness (1-7): Does the response offer truthful and detailed facts about the virtual character?

        ~
        
        [Evaluation Steps]
        
        1. Read through the interactions and identify the key points related to the character.
        
        2. Read through the responses of the AI assistant and compare them to the profile. Check if the responses are consistent with the character's profile, background, and known facts about the character.
        
        3. Check whether the responses provide detailed facts about the character or if they are generic responses that could apply to any character. Detailed responses are more factual and contribute positively to the score.
        
        4. Rate the performance of the AI on a scale of 1-7 for virtual knowledge correctness, where 1 is the lowest and 7 is the highest based on the Evaluation Criteria.
        
        ***
        
        ~
        
        First, write out in a step by step manner your reasoning about the criterion to be sure that your conclusion is correct. Avoid simply stating the correct answers at the outset. Then print the score on its own line corresponding to the correct answer. At the end, repeat just the selected score again by itself on a new line.\\
        \bottomrule
    \end{tabular}
    \caption{Prompt for ChatGPT to evaluate Virtual Knowledge.}
    \label{tab:judge_prompt_virtual}
\end{table*}

\begin{table*}[htbp]
    \scriptsize
    \centering
    \begin{tabular}{@{}p{\linewidth}@{}}
        \toprule
        \textbf{Prompt for Evaluation of Real Knowledge (RK)} \\
        \midrule
        You will be given responses written by an AI assistant mimicking the character \{agent\_name\}. Your task is to rate the performance of \{agent\_name\} using the specific criterion by following the evaluation steps. Below is the data:

        ~
        
        ***
        
        [Profile]
        
        \{agent\_context\}
        
        ~
        
        [Background]
        
        Location: \{loc\_time\}
        
        Status: \{status\}
        
        ***
        
        [Interactions]
        
        \{interactions\}
        
        ***
        
        [Evaluation Criterion]
        
        Real Knowledge Correctness (1-7): Is the response free from conflicts with the real-world knowledge?

        ~
        
        [Evaluation Steps]
        
        1. Read through the interactions and identify the key points related to the real-world knowledge.
        
        2. Read through the responses of the AI assistant and compare them to real-world knowledge. Check if the responses align with facts, events, and information that are generally accepted as true in the real world.
        
        3. Evaluate whether the responses demonstrate a clear understanding of real-world concepts and provide accurate information. Look for any instances where the AI may have provided information that contradicts established facts or where it lacks accuracy in representing real-world knowledge.
        
        4. Rate the performance of the AI on a scale of 1-7 for real knowledge correctness, where 1 is the lowest and 7 is the highest based on the Evaluation Criterion. Assign a higher score for responses that consistently align with real-world knowledge and a lower score for those with noticeable discrepancies or inaccuracies.
        
        ***
        
        ~
        
        First, write out in a step by step manner your reasoning about the criterion to be sure that your conclusion is correct. Avoid simply stating the correct answers at the outset. Then print the score on its own line corresponding to the correct answer. At the end, repeat just the selected score again by itself on a new line.\\
        \bottomrule
    \end{tabular}
    \caption{Prompt for ChatGPT to evaluate Real Knowledge.}
    \label{tab:judge_prompt_real}
\end{table*}

\begin{table*}[htbp]
    \scriptsize
    \centering
    \begin{tabular}{@{}p{\linewidth}@{}}
        \toprule
        \textbf{Prompt for Evaluation of Hallucinatory Knowledge (HK)} \\
        \midrule
        You will be given responses written by an AI assistant mimicking the character \{agent\_name\}. Your task is to rate the performance of \{agent\_name\} using the specific criterion by following the evaluation steps. Below is the data:
        
        ~
        
        ***
        
        [Profile]
        
        \{agent\_context\}
        
        ~
        
        [Background]
        
        Location: \{loc\_time\}
        
        Status: \{status\}
        
        ***
        
        [Interactions]
        
        \{interactions\}
        
        ***
        
        [Evaluation Criterion]
        
        Avoiding Hallucination (1-7): Does the response integrate real-world knowledge with knowledge about virtual characters?
    
        ~
        
        [Evaluation Steps]
        
        1. Read through the interactions and find the evidences about combining real-world knowledge and virtual characters knowledge.
        
        2. Look for clear distinctions between real-world information and details related to virtual characters.
        
        3. Compare the evidences to the profile. Check if the evidence combines real-world and virtual knowledge, leading to conflicts with the character's knowledge scope. If some evidences contradicts to the character's identity, given a lower score. Otherwise, assign a higher score.
        
        4. Rate the performance of the AI on a scale of 1-7 for Avoiding Hallucination, where 1 is the lowest and 7 is the highest based on the Evaluation Criteria.
        
        ***
        
        ~
        
        First, write out in a step by step manner your reasoning about the criterion to be sure that your conclusion is correct. Avoid simply stating the correct answers at the outset. Then print the score on its own line corresponding to the correct answer. At the end, repeat just the selected score again by itself on a new line.\\
        \bottomrule
    \end{tabular}
    \caption{Prompt for ChatGPT to evaluate Hallucinatory Knowledge.}
    \label{tab:judge_prompt_hallucinatory}
\end{table*}

\begin{table*}[htbp]
    \scriptsize
    \centering
    \begin{tabular}{@{}p{\linewidth}@{}}
        \toprule
        \textbf{Prompt for Evaluation of Stability (Stab.)} \\
        \midrule
        You will be given responses written by an AI assistant mimicking the character \{agent\_name\}. Your task is to rate the performance of \{agent\_name\} using the specific criterion by following the evaluation steps. Below is the data:

        ~
        
        ***
        
        [Profile]
        
        \{agent\_context\}
        
        ~
        
        [Background]
        
        Location: \{loc\_time\}
        
        Status: \{status\}
        
        ***
        
        [Interactions]
        
        \{interactions\}
        
        ***
        
        [Evaluation Criterion]
        
        Long-term Acting (1-7): Is the assistant maintain a good performance over the long interactions?

        ~
        
        [Evaluation Steps]
        
        1. Read through the given profile and background information to familiarize yourself with the context and details of the AI assistant named \{agent\_name\}.
        
        2. Review the interactions provided to see how \{agent\_name\} responds to various prompts and queries. And evaluate the performance of acting query by query that whether the response reflects the personalities, speaking styles, and values of the character. Assign score for each turn.
        
        3. Based on the above assigned scores, does \{agent\_name\} keep acting like character in the long-term? Evaluate the overall performance of the whole conversation based on the score for each turn.
        
        4. Rate the stability of \{agent\_name\} on a scale of 1 to 7, with 1 being very poor and 7 being excellent.
        
        ***
        
        ~
        
        First, write out in a step by step manner your reasoning about the criterion to be sure that your conclusion is correct. Avoid simply stating the correct answers at the outset. Then print the score on its own line corresponding to the correct answer. At the end, repeat just the selected score again by itself on a new line.\\
        \bottomrule
    \end{tabular}
    \caption{Prompt for ChatGPT to evaluate Stability.}
    \label{tab:judge_prompt_stability}
\end{table*}

\begin{table*}[htbp]
    \scriptsize
    \centering
    \begin{tabular}{@{}p{\linewidth}@{}}
        \toprule
        \textbf{Prompt for Evaluation of Relevance (Rel.)} \\
        \midrule
        You will be given responses written by an AI assistant mimicking the character \{agent\_name\}. Your task is to rate the performance of \{agent\_name\} using the specific criterion by following the evaluation steps. Below is the data:

        ~
        
        ***
        
        [Profile]
        
        \{agent\_context\}
        
        ~
        
        [Background]
        
        Location: \{loc\_time\}
        
        Status: \{status\}
        
        ***
        
        [Interactions]
        
        \{interactions\}
        
        ***
        
        [Evaluation Criterion]
        
        Relevance (1-7): Is the response relevant to the topic of given question in interactions?

        ~
        
        [Evaluation Steps]
        
        1. Read through the interactions and pinpoint the main topic of given question.
        
        2. Read through the responses of the AI assistant and compare them to the topic. Check if the responses are consistent with the topic of the given question.
        
        3. Evaluate whether the responses demonstrate a clear understanding of the topic. Look for any instances of conflicting information or inaccuracies.
        
        4. Rate the performance of the AI on a scale of 1-7 for Relevance, where 1 is the lowest and 7 is the highest based on the Evaluation Criterion. Assign a higher score for responses that consistently align with the topic of the question and a lower score for those with noticeable discrepancies or inaccuracies.
        
        ***
        
        ~
        
        First, write out in a step by step manner your reasoning about the criterion to be sure that your conclusion is correct. Avoid simply stating the correct answers at the outset. Then print the score on its own line corresponding to the correct answer. At the end, repeat just the selected score again by itself on a new line.\\
        \bottomrule
    \end{tabular}
    \caption{Prompt for ChatGPT to evaluate Relevance.}
    \label{tab:judge_prompt_relevance}
\end{table*}

\begin{table*}[htbp]
    \scriptsize
    \centering
    \begin{tabular}{@{}p{\linewidth}@{}}
        \toprule
        \textbf{Prompt for Transfer (Trans.)} \\
        \midrule
        You will be given responses written by an AI assistant mimicking multiple characters \{agent\_name\}. Your task is to rate the performance of \{agent\_name\} using the specific criterion by following the evaluation steps. Below is the data:

        ~
        
        ***
        
        [Profile]
        
        \{agent\_context\}
        
        ~
        
        [Background]
        
        Location: \{loc\_time\}
        
        Status: \{status\}
        
        ***
        
        [Interactions]
        
        \{interactions\}
        
        ***
        
        [Evaluation Criterion]
        
        Transfer (1-7): Does the AI assistant seamlessly transition between different roles, maintaining consistency and authenticity in each character portrayal?

        ~
        
        [Evaluation Steps]
        
        1. Review the interactions between the AI assistant and the user, focusing on instances where the AI switches between different characters.
        
        2. Assess the transitions between roles to determine if the AI maintains consistency and authenticity in each character portrayal. Look for smooth shifts in dialogue style, language usage, and personality traits that align with the characteristics of each character.
        
        3. Evaluate whether the AI effectively captures the essence of each character, ensuring that their responses reflect their historical or fictional background, personality traits, and mannerisms.
        
        4. Rate the performance of the AI on a scale of 1-7 for Transfer, where 1 represents a poor transition with inconsistencies in character portrayal, and 7 represents seamless transitions with each character authentically represented throughout the conversation. Assign a higher score for responses that demonstrate clear distinctions between characters and maintain consistency in their portrayal and a lower score for instances of ambiguity or inconsistency in character transitions.
        
        ***
        
        ~
        
        First, write out in a step by step manner your reasoning about the criterion to be sure that your conclusion is correct. Avoid simply stating the correct answers at the outset. Then print the score on its own line corresponding to the correct answer. At the end, repeat just the selected score again by itself on a new line.\\
        \bottomrule
    \end{tabular}
    \caption{Prompt for ChatGPT to evaluate Transfer.}
    \label{tab:judge_prompt_transfer}
\end{table*}
\end{document}